\documentclass[10pt,twocolumn,letterpaper]{article}

\usepackage{cvpr}
\usepackage{times}
\usepackage{epsfig}
\usepackage{graphicx}
\usepackage{amsmath}
\usepackage{amssymb}
\usepackage{multirow}
\usepackage{subfig}
\usepackage{tabularx}
\usepackage{comment}
\usepackage{sidecap}
\usepackage{dcolumn}
\usepackage{xcolor}

\definecolor{mod}{rgb}{0,0,0}

\usepackage[pagebackref=false,breaklinks=true,letterpaper=true,colorlinks,bookmarks=false]{hyperref}

\cvprfinalcopy 

\def\cvprPaperID{334} 

\DeclareMathOperator*{\argmin}{argmin}

\ifcvprfinal\fi
\begin{document}

\title{Anticipating Visual Representations from Unlabeled Video}

\author{Carl Vondrick \hspace{1em} Hamed Pirsiavash$\dagger$ \hspace{1em} Antonio Torralba\\
Massachusetts Institute of Technology \hspace{1em} $\dagger$University of Maryland, Baltimore County\\
\texttt{\{vondrick,torralba\}@mit.edu} \hspace{1em} \texttt{hpirsiav@umbc.edu}
}

\maketitle

\begin{abstract}

Anticipating actions and objects before they start or appear is a difficult
problem in computer vision with several real-world applications. 
This task is challenging partly because it
requires leveraging extensive knowledge of the world that is difficult to write
down. We believe that a promising resource for efficiently learning this
knowledge is through readily available unlabeled video. We
present a framework that capitalizes on temporal structure in
unlabeled video to learn to anticipate human actions and objects. The key idea
behind our approach is that we can train deep networks to predict the visual
representation of images in the future. Visual representations are a promising
prediction target because they encode images at a higher semantic level than
pixels yet are automatic to compute. We then apply recognition
algorithms on our predicted representation to anticipate objects and actions.
We experimentally validate this idea on two datasets, anticipating
actions one second in the future and objects five seconds in the future.

\end{abstract}

\section{Introduction}

\begin{figure}[tb]
\centering
\includegraphics[width=\linewidth]{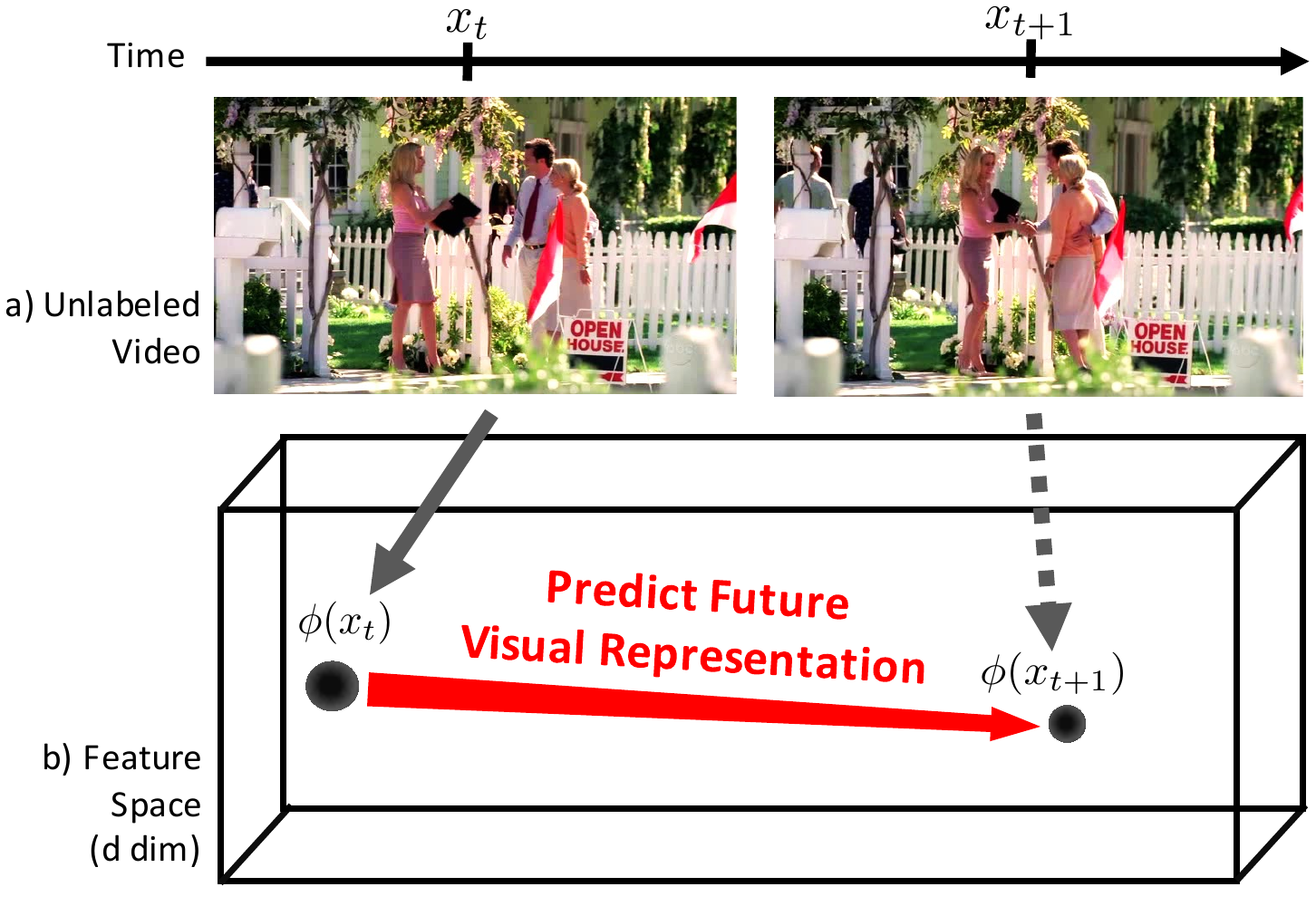} 
\vspace{-1em}
\caption{\textbf{Predicting Representations:} In this paper, we explore how to anticipate human actions and objects
by learning from unlabeled video. We propose to anticipate the visual representation of frames in the future. We can apply
recognition algorithms on the predicted representation to forecast actions and objects.\vspace{-1em}}
\label{fig:teaser}
\end{figure}

What action will the man do next in Figure \ref{fig:teaser} (left)?
A key problem in computer vision is to create machines that anticipate actions and objects 
in the future, before they appear or start.  This predictive capability would
enable several real-world applications. For example, robots can use predictions of human
actions to make better plans and interactions \cite{koppula2013anticipating}.  Recommendation systems can suggest products or
services based on what they anticipate a person will do. Predictive models can
also find abnormal situations in surveillance videos, and alert emergency
responders.

Unfortunately, developing an algorithm to anticipate the future is challenging.
Humans can rely on extensive knowledge accumulated over their lifetime to infer that the
man will soon shake hands in Figure \ref{fig:teaser}. How do we give machines access to this
knowledge?

We believe that a promising resource to train predictive models are abundantly
available unlabeled videos.  Although lacking ground truth annotations, they
are attractive for prediction because they are economical to obtain at
massive scales yet still contain rich signals. 
Videos come with the temporal ordering of frames ``for
free'', which is a valuable asset for forecasting. 

However, how to leverage unlabeled video to anticipate high-level concepts
is unclear.  Pioneering work in computer vision has capitalized on unlabeled videos
before to visualize the future
\cite{ranzato2014video,unsupervised,walker2014patch} and predict motions
\cite{pintea2014deja,walker2015dense,yuen2010data}. Unfortunately, these
self-supervised approaches are not straightforward to apply for anticipating semantics
because, unlike pixels or motions, concepts are not readily accessible in
unlabeled video. Methods that anticipate concepts have typically required supervision \cite{kitani2012activity,lan2014hierarchical,huang2014action}, which is expensive to scale. 

In this paper, we propose a method to anticipate concepts in the future
by learning from unlabeled video.  Recent progress in computer vision has built
rich visual representations
\cite{donahue2013decaf,razavian2014cnn,zha2015exploiting}.  Rather than predict
pixels or depend on supervision, our main idea is to forecast visual
representations of future frames.  Since these representations contain signals
sufficient to recognize concepts in the present, we then use recognition
algorithms on the forecasted representation to anticipate a future concept.
Representations have the advantage that they both a) capture the semantic
information that we want to forecast and b) scale to unlabeled videos because
they are automatic to compute. Moreover, representations may be easier to
predict than pixels because distance metrics in this space empirically tend to
be more robust \cite{donahue2013decaf,krizhevsky2012imagenet}.

Since we can economically acquire a large amounts of unlabeled video, we create our
prediction models with deep networks, which are attractive for this problem because their capacity can
grow with the size of data available and are trained efficiently with
large-scale optimization algorithms. In our experiments, we downloaded
$600$ hours of unlabeled video from the web
and trained our network to forecast representations $1$ to $5$ seconds in the future.
We then forecast both actions and objects by applying recognition algorithms
on top of our predicted representations. 
We evaluate this idea on two datasets of human actions in television shows
\cite{patron2010high} and egocentric videos for activities of daily living
\cite{pirsiavash2012detecting}. 
Although we are still far from human performance on these tasks, our
experiments suggest that learning to forecast representations with unlabeled videos may help
machines anticipate some objects and actions.

The primary contribution of this paper is developing a method to 
leverage unlabeled video for forecasting high-level concepts.
In section 2, we first review related work. In section 3, we then
present our deep network to predict visual representations in the future. Since
the future can be uncertain, we extend our network
architecture to produce multiple predictions. In section 4, we show
experiments to forecast both actions and objects. We plan to make our
trained models and code publicly available.

\section{Related Work}

The problem of predicting the future in images and videos has received growing
interest in the computer vision community, which our work builds upon:

\textbf{Prediction with Unlabeled Videos:}  Perhaps the ideas most similar to this paper are the ones that
capitalize on the wide availability of big video collections. In
early work, Yuen and Torralba \cite{yuen2010data} propose to predict motion in a single image by
transferring motion cues from visually similar videos in a large database. Building on the rich potential of large video collections,
Walker et al.\ \cite{walker2014patch} demonstrate a compelling data-driven approach that animates the
trajectory of objects from a single frame.
Ranzato et
al.\ \cite{ranzato2014video} and Srivastava et al.\ \cite{unsupervised} also
learn predictive models from large unlabeled video datasets to predict pixels
in the future. 
In this paper, we also use large video collections. However, unlike previous work
that predicts low-level pixels or motions, we develop a system to predict
high-level concepts such as objects and actions by learning from unlabeled video.

\begin{figure*}
\centering
\includegraphics[width=0.8\linewidth]{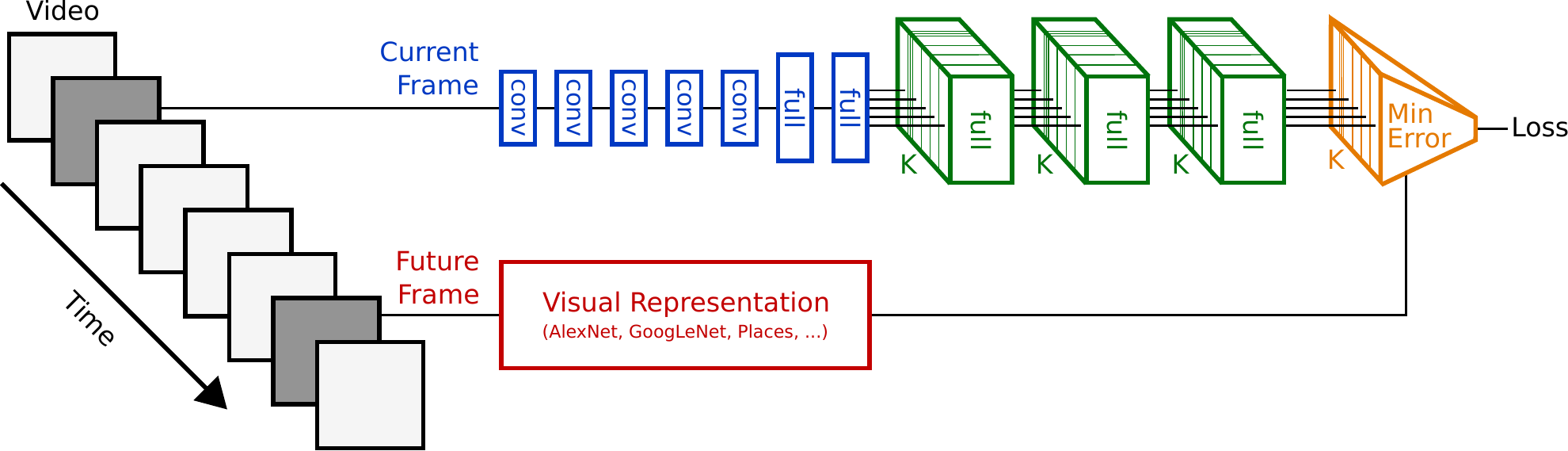}

\caption{\textbf{Network Diagram:} We visualize the network architecture we use
in our experiments. During training, the network uses videos to learn to
predict the representation of frames in the future. Since predicting the future
is a multi-modal problem, our network predicts $K$ future representations. Blue
layers are the same for each output while green layers are separate for the $K$
outputs.  During inference, we only input the current frame, and the network
estimates $K$ representations for the future. Please see section
\ref{sec:method} for full details.}

\label{fig:network}
\end{figure*}

\textbf{Predicting Actions:} There have been some promising works on
predicting future action categories. Lan et al. \cite{lan2014hierarchical}
propose a hierarchical representation to predict future actions in the wild.
Ryoo \cite{ryoo2011human} and Hoai and De la Torre \cite{hoai2014max} propose
models to predict actions in early stages.  Vu et al.\ in
\cite{vu2014predicting} learn scene affordance to predict what actions can
happen in a static scene.  Pei et al.\ \cite{pei2011parsing} and Xie et al.\
\cite{xie2013inferring} infer people's intention in performing actions which is
a good clue for predicting future actions. We are different from these
approaches because we use large-scale unlabeled data to predict a rich visual
representation in the future, and apply it towards anticipating both actions
and objects.

\textbf{Predicting Human Paths:} There have been several works 
that predict the future by reasoning about scene semantics with encouraging success.
Kitani et al. \cite{kitani2012activity} use concept detectors to predict the
possible trajectories a person may take in surveillance applications.
Lezema et al.\ 
\cite{lezama2011track}, Gong et al.\ \cite{gong2011multi} and Kooij et al.\ \cite{kooij2014context} also predict the
possible future path for people in the scene. Koppula and Saxena \cite{koppula2013anticipating}
anticipate the action movements a person may take in a human robot interaction scenario using RGB-D sensors.
Our approach extends these efforts by predicting human actions and objects.

\textbf{Predicting Motions:}
One fundamental component of prediction is predicting short motions, and there have been
some investigations towards this.  Pickup et al.\ in \cite{arrowoftime}
implicitly model causality to understand what should happen before what in a
video. Fouhey and Zitnick \cite{fouhey2014predicting} learn from abstract scenes to predict what
objects may move together. Lampert \cite{lampert2015predicting} predicts the future state of a probability distribution, and applies it towards predicting classifiers
adapted to future domains. Pintea et al.\ \cite{pintea2014deja} predict the optical flow from
single images by predicting how pixels are going to move in future.
We are
hoping that our model learns to extrapolate these motions automatically in the visual representation, which is helpful if we want
to perform recognition in the future rather than rendering it in pixel space. 


\textbf{Big Visual Data:}
We build upon work that leverages a large amount of visual data
readily available online. Torralba et al.\ \cite{torralba200880} use millions
of Internet images to build object and scene recognition systems. 
Chen et al.\ \cite{chen2013neil} and Divvala et al.\ \cite{divvalalearning}
build object recognition systems that have access to common sense by mining
visual data from the web.  Doersch et al. \cite{doersch2012makes} use large
repositories of images from the web to tease apart visually distinctive
elements of places. 
Kim and Xing \cite{kim2014reconstructing} learn to reconstruct story lines in personal photos,
and recommend future photos.
Zhou et al.\ \cite{zhoulearning} train convolutional neural
networks on a massive number of scene images to improve scene recognition
accuracy. In our work, we also propose to mine information from visual media on
the web, however we do it for videos with the goal of learning a model to
anticipate semantic concepts.

\textbf{Unsupervised Learning in Vision:} To handle large-scale data, there
have been some efforts to create unsupervised learning systems for vision.
Ramanan et al.\ \cite{ramanan2007leveraging} uses temporal relationships in
videos to build datasets of human faces. Ikizler-Cinbis et al.\
\cite{ikizler2009learning} propose to use images from the web to learn and
annotate actions in videos without supervision.  Le et al.\
\cite{le2013building} show that machines can learn to recognize both human and cat faces by watching an enormous amount of YouTube
videos. Chen and
Grauman \cite{chen2013watching} propose a method to discover new human actions
by only analyzing unlabeled videos, and Mobahi et al.\ \cite{mobahi2009deep}
similarly discover objects. This
paper also proposes to use unlabeled data, but we use unlabeled video to learn to predict visual representations.

\textbf{Representation Learning:} Recent work has explored how to learn visual
representations, for example with images \cite{doersch2015unsupervised} or
videos \cite{wang2015unsupervised}. Our work is different because we do not seek to
learn a visual representation. Rather, our goal is to anticipate the visual
representation in the future. Moreover, our approach is general, and in
principle could predict any representation.


%
%




\section{Anticipating  Visual Representations}
\label{sec:method}

Rather than predicting pixels (which may be more difficult) 
or anticipating labeled categories (which requires supervision), our idea
is to use unlabeled video to learn to predict the visual representation
in the future. We can then apply recognition algorithms (such as object or action classifiers)
on the predicted future representation to anticipate a high-level concept.  In this section,
we explain our approach.

\subsection{Self-supervised Learning}


Given a video frame $x_t^i$ at time $t$ from video $i$, our goal is to predict the visual
representation for the future frame $x_{t+\Delta}^i$. Let $\phi(x_{t+\Delta}^i)$ be
the representation in the future. Using videos as training data, we wish to estimate a function $g(x_t^i)$ that
closely predicts $\phi(x_{t+\Delta}^i)$:
\begin{align}
\omega^* = \argmin_\omega \sum_{i,t} \lVert g\left(x_t^i; \omega\right) - \phi\left(x_{t+\Delta}^i\right) \rVert_2^2
\label{eqn:loss}
\end{align}
where our prediction function $g(\cdot)$ is parameterized by $\omega$.

Our method is general to most visual representations, however we focus on predicting the last hidden layer  ($\texttt{fc7}$) of AlexNet
\cite{krizhevsky2012imagenet}. We chose this layer because it empirically
obtains state-of-the-art performance on several image
\cite{razavian2014cnn,donahue2013decaf} and video \cite{zha2015exploiting}
recognition tasks.

%

\subsection{Deep Regression Network}

Since we do not require data to be labeled for learning, we can collect
large amounts of training data. We propose to use deep regression networks for predicting representations because
their model complexity can expand to harness the amount of data available and can be trained with large scale
data efficiently with stochastic gradient descent.

Our network architecture is five convolutional layers followed by five
fully connected layers. The last layer is the output vector, which makes the
prediction for the future representation.  In training, we use a Euclidean
loss to minimize the distance between our predictions $g(x_t)$ and the 
representation of the future frame $\phi(x_{t+\Delta})$.

Our choice of architecture is motivated by the
successes of the AlexNet architecture for visual recognition
\cite{krizhevsky2012imagenet,zhoulearning}. However, 
our architecture differs by having a regression loss function 
and three more fully connected layers. 


\begin{figure}[t]
\includegraphics[width=1\linewidth]{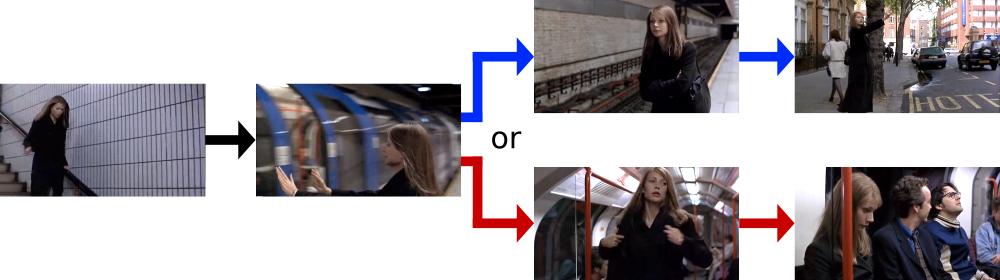}
\caption{\textbf{Multiple Futures:} Since the future can be uncertain, our model
anticipates multiple possibilities.}
\label{fig:multi}
\end{figure}

\subsection{Multi-Modal Outputs}

Given an image, there can be multiple plausible futures, illustrated in
Figure \ref{fig:multi}. We wish to handle the multi-modal nature of this problem
for two reasons. Firstly, when there are multi-modal outputs, the optimal least
squares solution for regression is to produce the mean of the modes. This is
undesirable because the mean may either be unlikely or off the manifold of
representations. Secondly, reasoning about uncertain
outcomes can be important for some applications of future prediction.


We therefore extend deep regression networks to produce multiple outputs.
Suppose that there are $K$ possible output vectors for one input frame.  We can 
support multiple outputs by training a mixture of $K$ networks, where
each mixture is trained to predict one of the modes in the future. 
Given input $x_t^i$, network $k$ will produce one of the outputs $g_k(x_t^i)$.

\subsection{Learning}


To train multiple regression networks, we must address two challenges. Firstly, videos only show
one of the possible futures (videos like Figure \ref{fig:multi} are rare).
Secondly, we do not know to \emph{which} of the $K$ mixtures each frame belongs.
We overcome both problems by treating the mixture assignment for a
frame as latent.

Let $z_t^i \in \{1, \ldots, K\}$ be a latent variable indicating
this assignment for frame $t$ in video $i$.
We first initialize $z$ uniformly at random. Then, we alternate
between two steps. First, we solve for the network weights $w$ end-to-end using backpropagation
assuming $z$ is fixed:
\begin{align}
\omega^* = \argmin_\omega \sum_{i,t} \left|\left| g_{z_t^i}\left(x_t^i; \omega\right) - \phi\left(x_{t+\Delta}^i\right) \right|\right|_2^2
\end{align}
Then, we re-estimate $z$ using the new network weights:
\begin{align}
z_t^i = \argmin_{k \in \{1, \ldots, K\}} \left|\left| g_k\left(x_t^i; \omega\right) - \phi\left(x_{t+\Delta}^i\right) \right|\right|_2^2
\end{align}
We alternate between these two steps several times, a process
that typically takes two days. We learn $w$ with warm starting, and let it train
for a fixed number of iterations before updating $z$. We illustrate this
network in Figure \ref{fig:network}.

Although we train our network offline in our experiments, we note our network can be
also be trained online with streaming videos. Online learning is attractive because the network
can continuously learn how to anticipate the future without storing frames. Additionally, the model can adapt
in real time to the environment, which may be useful in some applications.

\subsection{Predicting Categories}

Since our network uses unlabeled videos to predict a 
representation in the future, we need a way to attach semantic category labels to it. To do this, we use a relatively small set of labeled examples
from the target task to indicate the category of interest. As the representation that we predict is
the same that is used by state-of-the-art recognition systems, we can 
apply standard recognition algorithms to the predicted representation in order
to forecast a category. 

We explore two strategies for using recognition algorithms on the predicted
representations. The first strategy uses a visual classifier trained on the
standard features (we use \texttt{fc7}) from frames containing the category of interest, but applies
it on a predicted representation. The second strategy trains the
visual classifier on the predicted representations as well. The second strategy
has the advantage that it can adapt to structured errors in the regression.

During inference, our model will predict multiple representations of the
future. By applying category classifiers to each predicted representation, we
will obtain a distribution for how likely categories are to happen in
each future representation. We marginalize over these distributions to obtain
the most likely category in the future.

\subsection{Implementation}

Our network architecture consists of 5 convolutional layers followed by 5 fully
connected layers. We use ReLU nonlinear activations throughout the network. The
convolutional part follows the AlexNet architecture, and we refer readers to
\cite{krizhevsky2012imagenet} for complete details. After the convolutional
layers, we have 5 fully connected layers each with $4096$ hidden units. 


The $K$ networks (for each output) can either be disjoint or share parameters between them.
In our experiments, we opted to use the following sharing strategy in order to reduce the number
of free parameters. For the five convolutional layers and first two hidden layers, we
tie them across each mixture. For the last three fully connected layers, we
interleave hidden units: we randomly commit each hidden unit to a
network with probability $p = \frac{1}{2}$, which controls the amount of
sharing between networks.  We do this assignment once, and do not
change it during learning.

We trained the networks jointly with stochastic gradient descent. We used a
Tesla K40 GPU and implemented the network in Caffe \cite{jia2014caffe}.
We modified the learning procedure to
handle latent variables.  We
initialized the first seven layers of the network with the Places-CNN network
weights \cite{zhoulearning}, and the remaining layers with Gaussian white noise
 and the biases to a constant. During
learning, we also used dropout \cite{srivastava2014dropout} with a dropout
ratio of $\frac{1}{2}$ on every fully connected layer.  We used a fixed
learning rate of $.001$ and momentum term of $0.9$.

\begin{figure}
\centering
\includegraphics[width=\linewidth]{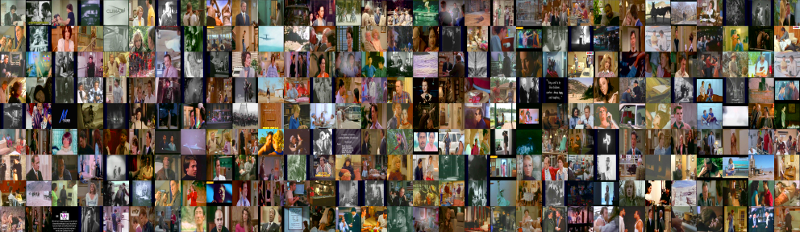}
\caption{\textbf{Unlabeled Videos:} We collected more than 600 hours of unlabeled video from YouTube. We show a sample of the frames above. We use this data to train deep networks that predict visual representations in the future.}
\label{fig:repository}
\end{figure}

\section{Experiments}

In this section, we experiment with how well actions and objects can be
forecasted using the predicted representations. We show results for forecasting
basic human actions one second before they start, and anticipating household
objects five seconds before they appear.


\subsection{Unlabeled Repository}

In order to train our network to predict features, we leverage a large amount
of unlabeled video.  We experimented with two sources of unlabeled videos:

\textbf{Television Shows:} We downloaded over $600$ hours of publicly available
television shows from YouTube. To pick the set of television shows, we used the
top shows according to Google. The videos we downloaded generally consist of
people performing a large variety of everyday actions, such as eating or
driving, as well as interactions with objects and other people.  We show a few
example frames of these videos in Figure \ref{fig:repository}. We use this
repository in most of our experiments.  Since we test on different datasets,
one concern is that there may be videos in the repository that also appear in a
testing set. To check this, we queried for nearest neighbors between this
repository and all testing sets, and found no overlap.

\textbf{THUMOS:} We
also experimented with using videos from the THUMOS challenge \cite{THUMOS15},
which consists of 400 hours of video from the web. These videos tend to be tutorials and sports, which
has a different distribution from television shows. 
We only use THUMOS as a diagnostic dataset to quantify the performance of our method
when the training distribution is very different from the testing set.

\newcommand{\tbrow}[9]{ #1 &  #4 & #2 & #5 & #7 & #6 & #3 & #8 & #9 \\}
\newcommand{\stc}[2][c]{%
  \begin{tabular}[#1]{@{}c@{}}#2\end{tabular}}
\begin{table*}
\begin{tabular}{l | c | c c c c | c c c}
       &         & \multicolumn{4}{c|}{Regression} & \multicolumn{3}{c}{Classifier} \\
Method & Feature & Train Data & Method & Output & K & Frame & Data &  Method \\
\hline
\tbrow{ SVM Static }{ - }{ During }{ fc7 }{ - }{ 1 }{ - }{ RO }{ SVM }
\tbrow{ SVM }{ - }{ Before }{ fc7 }{ - }{ 1 }{ - }{ RO }{ SVM }
\tbrow{ MMED }{ - }{ Before }{ fc7 }{ - }{ 1 }{ - }{ RO }{ MMED }
\tbrow{ Nearest Neighbor }{ UV }{ Before }{ fc7 }{ 1-NN }{ 1 }{ fc7 }{ RI }{ SVM }
\tbrow{ Nearest Neighbor Adapted }{ UV }{ Before }{ fc7 }{ 1-NN }{ 1 }{ fc7 }{ RO }{ SVM }
\tbrow{ Linear }{ UV }{ Before }{ fc7 }{ Linear }{ 1 }{ fc7 }{ RI }{ SVM }
\tbrow{ Linear Adapted }{ UV }{ Before }{ fc7 }{ Linear }{ 1 }{ fc7 }{ RO }{ SVM }
\tbrow{ Deep K=1 }{ UV }{ Before }{ RGB }{ CNN }{ 1 }{ fc7 }{ fc7 of RI }{ SVM }
\tbrow{ Deep K=1 Adapted }{ UV }{ Before }{ RGB }{ CNN }{ 1 }{ fc7 }{ RO }{ SVM }
\tbrow{ Deep K=3 }{ UV }{ Before }{ RGB }{ CNN }{ 3 }{ fc7 }{ fc7 of RI }{ SVM }
\tbrow{ Deep K=3 Adapted }{ UV }{ Before }{ RGB }{ CNN }{ 3 }{ fc7 }{ RO }{ SVM }
\tbrow{ Deep K=3 THUMOS  }{ THUMOS }{ Before }{ RGB }{ CNN }{ 3 }{ fc7 }{ fc7 of RI }{ SVM }
\tbrow{ Deep K=3 THUMOS Adapted }{ THUMOS }{ Before }{ RGB }{ CNN }{ 3 }{ fc7 }{ RO }{ SVM }
\tbrow{ Deep K=1 ActionBank Adapted }{ UV }{ Before }{ RGB }{ CNN }{ 1 }{ ActionBank }{ RO }{ SVM }
\tbrow{ Deep K=3 ActionBank Adapted }{ UV }{ Before }{ RGB }{ CNN }{ 3 }{ ActionBank }{ RO }{ SVM }
\end{tabular}
\caption{\textbf{Overview of Models:} We compare several different ways of training
models, and this table shows their different configurations. To train the regression (if any), we
specify which source of unlabeled videos we use (UV for our repository, or THUMOS), the method, the regression target output, and the number of outputs $K$. This is then fed into the classifier, which uses labeled data.  To train the classifier,
we specify which frame to train the classifier on (during action, or before action), the regression input (RI) or output (RO), 
and the classifier. During testing, the procedure is the same for all models.}
\label{tab:baselines} 
\end{table*}

\begin{table}
\centering
\begin{tabular}{l l}
Method & \multicolumn{1}{c}{Accuracy}  \\
\hline
Random & $25.0$ \\
SVM Static & $36.2 \pm 4.9$ \\
SVM & $35.8 \pm 4.3$ \\
\textcolor{mod}{MMED}  & $34.0 \pm 7.0$ \\ 
Nearest Neighbor  & $29.9 \pm 4.6$\\
Nearest Neighbor \cite{yuen2010data}, Adapted  & $34.9 \pm 4.7$\\
Linear & $32.8 \pm 6.1$ \\
Linear, Adapted & $34.1 \pm 4.8$ \\
\hline
Deep K=1, ActionBank \cite{sadanand2012action} & $34.0 \pm 6.1$ \\
Deep K=3, ActionBank \cite{sadanand2012action} & $35.7 \pm 6.2$ \\
Deep K=1 & $36.1 \pm 6.4$ \\
Deep K=1, Adapted & $40.0 \pm 4.9 $ \\
Deep K=3 & $35.4 \pm 5.2$ \\
\textbf{Deep K=3, Adapted} & $\textbf{43.3} \pm \textbf{4.7} $ \\
Deep K=3, THUMOS \cite{THUMOS15}, Off-the-shelf & \textcolor{mod}{$29.1 \pm 3.9$} \\
\textbf{Deep K=3, THUMOS \cite{THUMOS15}, Adapted } & $\textbf{43.6} \pm \textbf{4.8}$ \\
\hline
Human (single) & $71.7 \pm 4.2$ \\ 
Human (majority vote) & $85.8 \pm 1.6$ \\ 
%
\end{tabular}
\caption{\textbf{Action Prediction:} Classification accuracy for predicting actions one second before they begin given only a single frame. The standard deviation across cross-validation splits is next to the accuracy. 
}
\label{tab:actionresults}
\end{table}

\subsection{Baselines}

Our goal in this paper is to learn from unlabeled video to anticipate high-level concepts (specifically actions and objects) in the future.
Since our method uses minimal supervision to attach semantic meaning to the predicted representation,
we compare our model against baselines that use a similar level of supervision. See Table \ref{tab:baselines} for an overview of the methods we compare. 

\textbf{SVM:} One reasonable approach is to 
train a classifier on the frames before the action
starts to anticipate the category label in the future. This baseline is able to
adapt to contextual signals that may suggest the onset of an action. However,
since this method requires annotated videos, it does not capitalize on unlabeled video.

\textbf{MMED:} We can also extend the SVM to handle sequential data in order to make
early predictions. We use the code out-of-the-box provided by \cite{hoai2014max} for this
baseline.

\textbf{Nearest Neighbor:} Since we have a large unlabeled repository, one
reasonable approach is to search for the nearest neighbor, and use the
neighbor's future frame as the predicted representation, similar to
\cite{yuen2010data}.

\textbf{Linear:} Rather than training a deep network, we
can also train a linear regression on our unlabeled repository to predict
$\texttt{fc7}$ in the future. 


\textbf{Adaptation:} We also examine two
strategies for training the final classifier. One way is to train the
classifier on the ground truth regression targets, and test it on the inferred output of the regression. 
The second way is to adapt to the predictions by also training the
classifier on the inferred output of the regression. The latter can adapt to the errors
in the regression.


\subsection{Forecasting Actions}

\textbf{Dataset:} In order to evaluate our method for action forecasting, we require a labeled
testing set where a) actions are temporally annotated, b) we have access to
frames before the actions begin, and c) consist of everyday human actions (not sports). We use the TV Human Interactions dataset
\cite{patron2010high} because it satisfies these requirements. The
dataset consists of people performing four different actions (hand shake, high
five, hug, and kissing), with a total of $300$ videos.

\textbf{Setup:} We run our predictor on the frames before
the annotated action begins. We use the provided train-test splits with $25$-fold cross
validation. We evaluate classification accuracy (averaged across cross
validation folds) on making predictions one second before the action has
started.  To attach semantic meaning to our predicted representation, we use
the labeled examples from the training set in \cite{patron2010high}. As we make
multiple predictions, for evaluation purposes we consider a prediction to be
correct only if the ground truth action is the most likely prediction under our
model.

\begin{figure}
\centering
\includegraphics[width=\linewidth]{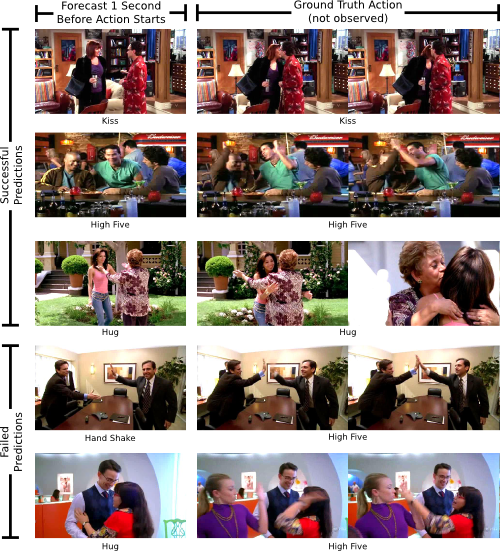}
\caption{\textbf{Example Action Forecasts:} We show some examples of our forecasts of actions one second before they begin. The left most column shows the frame before the action begins, and our forecast is below it. The right columns show the ground truth action. Note that our model does not observe the action frames during inference.}
\label{fig:actions-qual}
\end{figure}

%

\begin{figure}
\centering
\includegraphics[width=\linewidth]{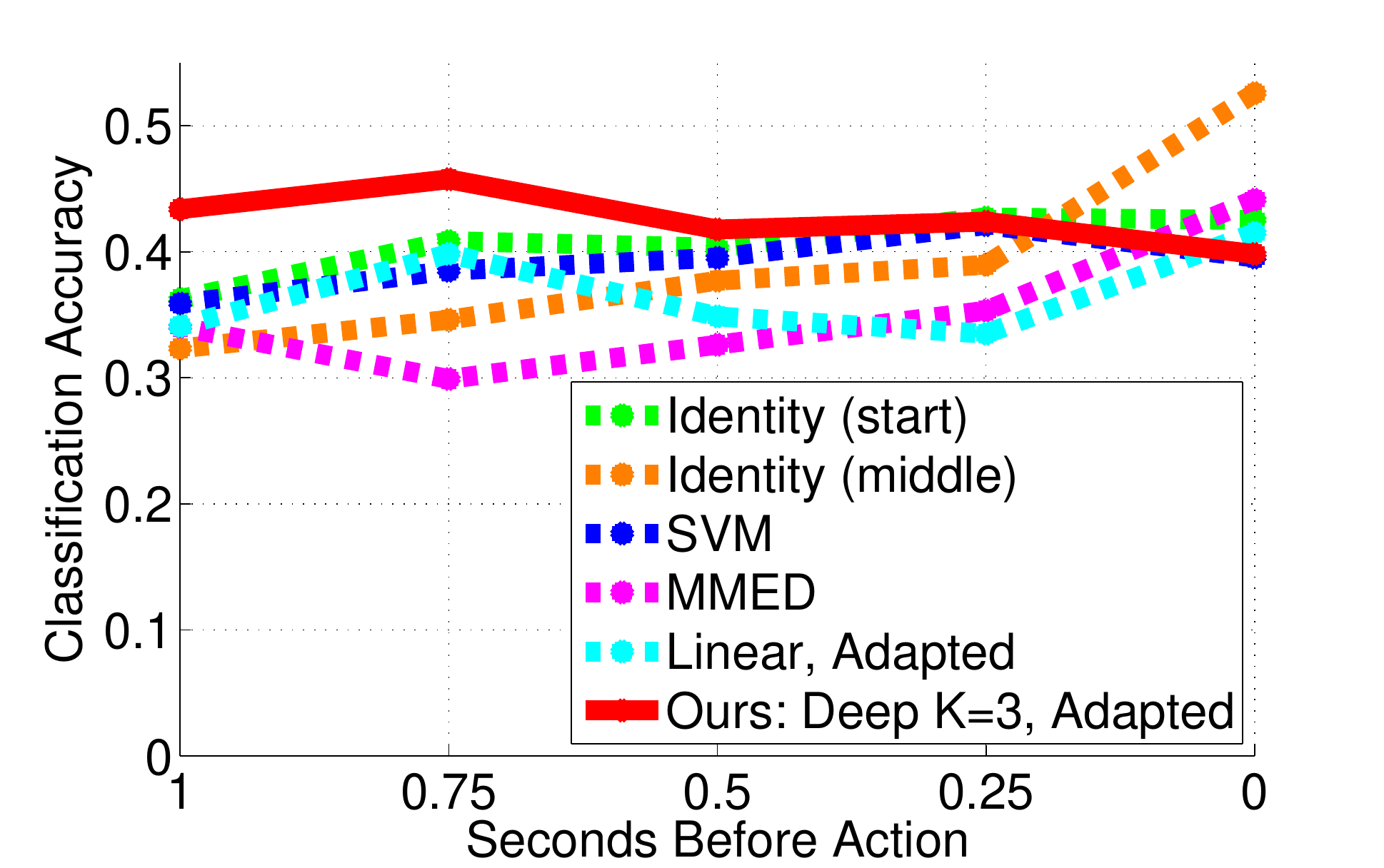}
\caption{\textcolor{mod}{\textbf{Performance vs $\Delta$:} We plot performance on forecasting actions versus number of frames before the action starts. Our model (red) performs better
when the time range is longer (left of plot). 
Note that, since our model takes days to train, we evaluate our model trained for one second, but evaluate on different time intervals. The baselines are trained for each time interval.}}
\label{fig:actions-delta}
\end{figure}

\begin{figure}
\includegraphics[width=\linewidth]{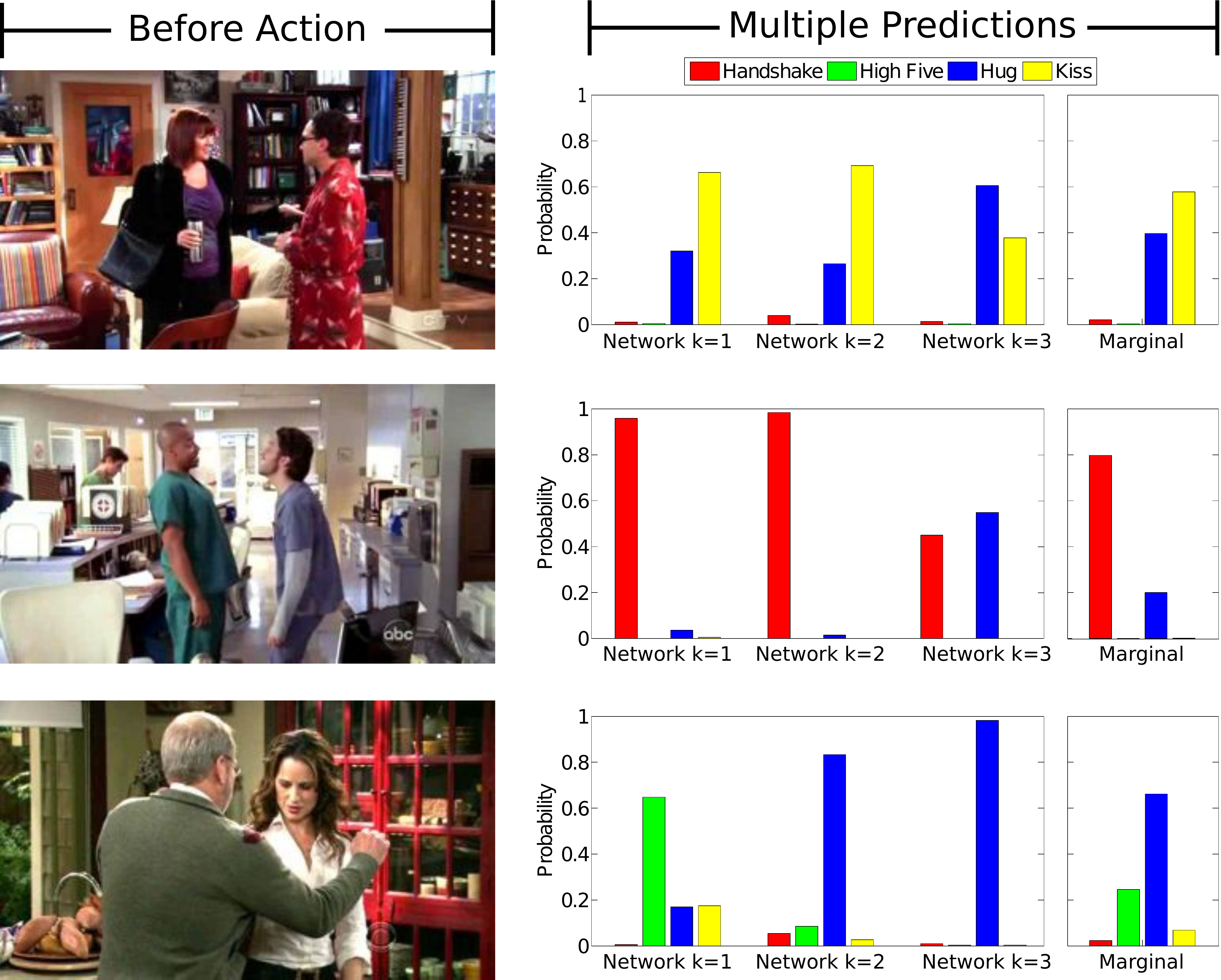}
\caption{\textcolor{mod}{\textbf{Multiple Predictions:} Given an input frame (left), our model predicts multiple representations in the future that can each be classified into actions (middle).
When the future is uncertain, each network can predict a different representation, allowing for multiple action forecasts. To obtain the most likely future action, we can marginalize
the distributions from each network (right).}}
\label{fig:actions-multi}
\end{figure}


\textbf{Results:}
Table \ref{tab:actionresults} shows the classification accuracy of different models for predicting the future action one second into
the future given only a single frame. Our results suggest that training deep
models to predict future representations with unlabeled videos may help
machines forecast actions, obtaining a
\textbf{relative gain of 19\%} over baselines.
We conjecture our network may obtain the stronger
performance partly because it can better predict the future \texttt{fc7}. The
mean Euclidean distance between our model's regressions and the actual future is about
$1789$, while regressing the identity transformation is about $1907$ and a
linear regression is worse, around $2328$.

\textbf{Human Performance:} To establish an upper expectation for the
performance on this task, we also had $12$ human volunteers study the training
sets and make predictions on our testing set. Human accuracy is good (an
average human correctly predicts 71\% of the time), but not perfect due to the
uncertain nature of the task. We believe humans are not perfect because the
future has inherent uncertainty, which motivates the need for models to make
multiple predictions. Interestingly, we can use the ``wisdom of the crowds'' to
ensemble the human predictions and evaluate the majority vote, which obtains
accuracy (85\%).

We also performed several experiments to breakdown the performance our method.
\textbf{Different Representations:} We
also tried to train a deep network to forecast ActionBank
\cite{sadanand2012action} in the future instead of $\texttt{fc7}$, which
performed worse. Representations are richer than action labels, which may
provide more constraints during learning that can help build more robust models
\cite{hinton2014distilling}. \textbf{Different Training Sets:} We also evaluated our network trained on videos
that are not television shows, such as sports and tutorials.
When we train our network with videos from THUMOS \cite{THUMOS15} instead of
our repository, we still obtain competitive performance, suggesting our method
may be robust to some dataset biases.  However, adaptation becomes more important
for THUMOS, likely because the classifier must adapt to the dataset bias. \textbf{Different Intervals:} We also evaluated our model varying the time before the
action starts in Figure
\ref{fig:actions-delta}. The relative gain of our method is often better as the prediction time frame increases.

%


\begin{figure}
\includegraphics[width=\linewidth]{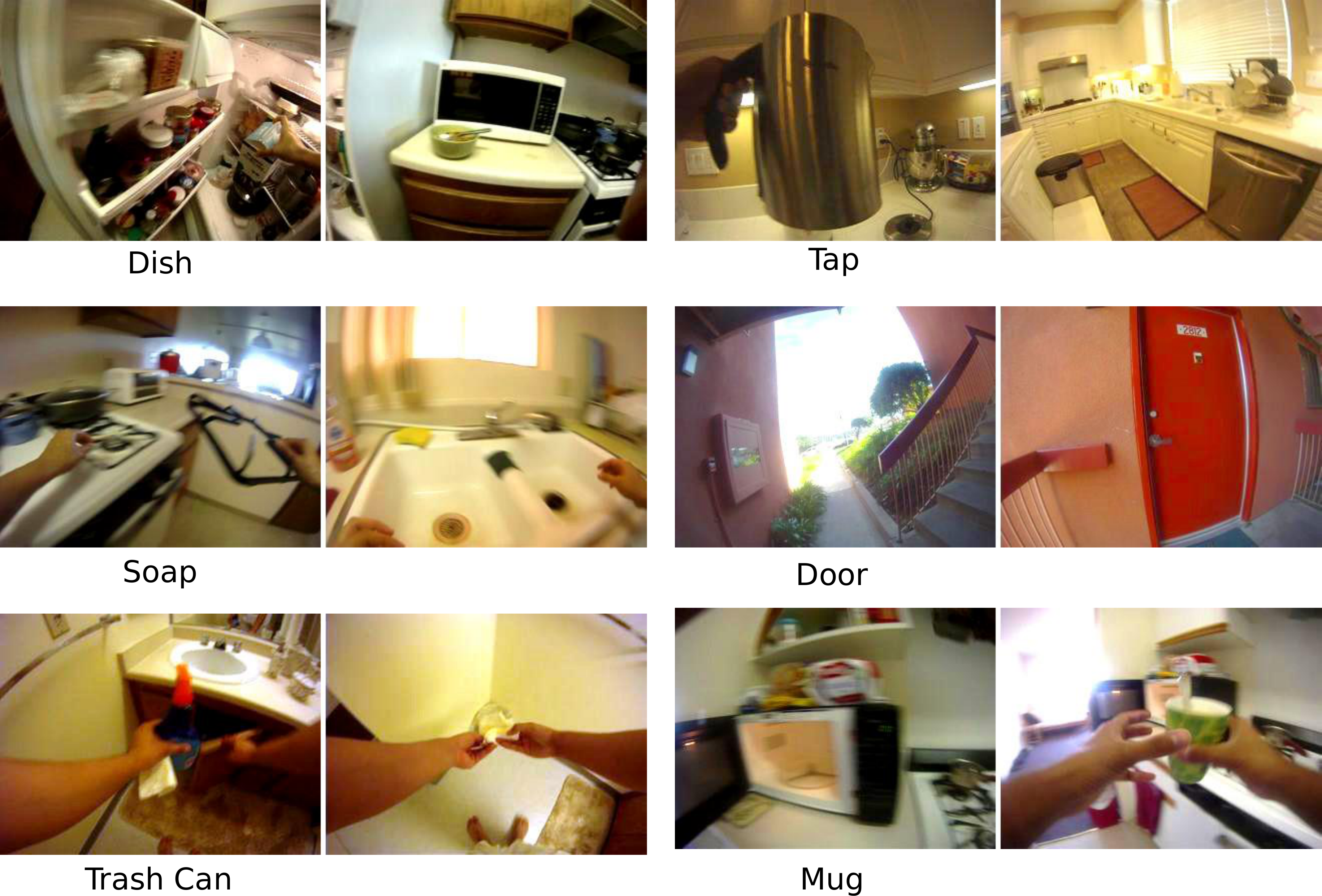}
\caption{\textbf{Example Object Forecasts:} We show examples of high scoring forecasts for objects. The left most frame is five seconds before the object appears.}
\label{fig:adl-qual}
\end{figure}

\newcolumntype{Y}{D..{2.1}}
\makeatletter
\newcolumntype{B}[3]{>{\boldmath\DC@{#1}{#2}{#3}}c<{\DC@end}}
\makeatother

\begin{table*}[tb]
\centering
\setlength{\tabcolsep}{.2em}
\begin{tabular}{l | Y | Y Y Y Y Y Y Y Y Y Y Y Y Y Y }
Method & \multicolumn{1}{c|}{Mean}& \multicolumn{1}{c}{dish} & \multicolumn{1}{c}{door} & \multicolumn{1}{c}{utensil} & \multicolumn{1}{c}{cup} & \multicolumn{1}{c}{oven} & \multicolumn{1}{c}{person} & \multicolumn{1}{c}{soap} & \multicolumn{1}{c}{tap} & \multicolumn{1}{c}{tbrush} & \multicolumn{1}{c}{tpaste} & \multicolumn{1}{c}{towel} & \multicolumn{1}{c}{trashc} & \multicolumn{1}{c}{tv} & \multicolumn{1}{c}{remote}  \\
\hline
Random & 1.2& 1.2 & 2.8 & 1.1 & 2.4 & 1.6 & 0.8 & 1.5 & 2.1 & 0.2 & 0.3 & 0.6 & 1.1 & 0.5 & 0.3 \\
SVM Static & 6.4& 2.6 & 15.4 & 2.9 & 5.0 & 9.4 & 6.9 & 11.5 & 17.6 & 1.6 & 1.0 & 1.5 & 6.0 & 2.0 & 5.9 \\
SVM & 5.3& 3.0 & 8.2 & 5.2 & 3.6 & 8.3 & 12.0 & 6.7 & 11.7 & 3.5 & 1.5 & 4.9 & 1.3 & 0.9 & 4.1 \\
Scene & 8.2& 3.3 & 18.5 & 5.6 & 3.6 & 18.2 & 10.8 & 9.2 & 6.8 & 8.0 & 8.1 & 5.1 & 5.7 & 2.0 & 10.3 \\
Scene, Adapted & 7.5& 4.6 & 9.1 & 6.1 & 5.7 & 15.4 & 13.9 & 5.0 & 15.7 & 13.6 & 3.7 & 6.5 & 2.4 & 1.8 & 1.7 \\
Linear & 6.3& 7.5 & 9.3 & 7.2 & 5.9 & 2.8 & 1.6 & 13.6 & 15.2 & 3.9 & 5.6 & 2.2 & 2.9 & 2.3 & 7.8 \\
Linear, Adapted & 5.3& 2.8 & 13.5 & 3.8 & 3.6 & 11.5 & 11.2 & 5.8 & 4.9 & 5.4 & 3.3 & 3.4 & 1.6 & 2.1 & 1.0 \\
\hline
Deep K=1& 9.1& 4.4 & 17.9 & 3.0 & 14.8 & 11.9 & 9.6 & 17.7 & 15.1 & 6.3 & 6.9 & 5.0 & 5.0 & 1.3 & 8.8 \\
Deep K=1, Adapted & 8.7& 3.5 & 11.0 & 9.0 & 6.5 & 16.7 & 16.4 & 8.4 & 22.2 & 12.4 & 7.4 & 5.0 & 1.9 & 1.6 & 0.5 \\
\textbf{Deep K=3} & \multicolumn{1}{B{.}{.}{1} |}{10.7}& 4.1 & 22.2 & 5.7 & 16.4 & 17.5 & 8.4 & 19.5 & 20.6 & 9.2 & 5.3 & 5.6 & 4.2 & 8.0 & 2.6 \\
\textbf{Deep K=3, Adapted}& \multicolumn{1}{B{.}{.}{1} |}{10.1} & 3.5 & 14.7 & 14.2 & 6.7 & 14.9 & 15.8 & 8.6 & 29.7 & 12.6 & 4.6 & 10.9 & 1.8 & 1.4 & 1.9 \\
\end{tabular}
\caption{\textbf{Object Prediction:} We show average precision for forecasting
objects five seconds before they appear in egocentric videos. For most
categories, our method improves prediction performance. 
The last column is the mean across all categories.}
\label{tab:adl}
\end{table*}

\textbf{Multiple Predictions:}
Since we learn a mixture of networks, our model can make diverse predictions
when the future is uncertain. To analyze this, Figure \ref{fig:actions-multi}
shows a scene and a distribution of possible future actions.  For example,
consider the first row where the man and woman are about to embrace, however
whether they will kiss or hug is ambiguous. In our model, two of the networks
predict a representation where kissing is the most likely future action, but
one network predicts a representation where the most likely action is hugging. The other rows
show similar scenarios.
Since performance drops when $K=1$,
modeling multiple outputs may be important both during learning and inference.

\textbf{Qualitative Results:}
We qualitatively show some of our predictions in Figure \ref{fig:actions-qual}. For example, in some cases our model correctly predicts that a man and
woman are about to kiss or hug or that men in a bar will high five.  The second to last row shows a comic scene where one
man is about to handshake and the other is about to high five, which our model
confuses. In the last row of Figure \ref{fig:actions-qual}, our model
incorrectly forecasts a hug because a third person unexpectedly enters the
scene.


\subsection{Forecasting Objects}

\textbf{Dataset:}
Since our method predicts a visual representation in the future,  we wish to
understand how well we can anticipate concepts other than actions.  We
experimented with forecasting objects in egocentric videos five seconds before
the object appears. We use the videos from Activities of the Daily Living
dataset \cite{pirsiavash2012detecting}, which is one of the largest 
datasets of egocentric videos from multiple people. Anticipating objects in this dataset
is challenging because even recognizing
objects in these videos is difficult \cite{pirsiavash2012detecting}.

\textbf{Setup:}
In order to train our deep network on egocentric videos, we reserved three
fourths of the dataset as our repository for self-supervised learning. We 
evaluate on the remaining one fourth videos, performing leave-one-out to learn
future object category labels. Since multiple objects can appear in a frame, we
evaluate the average precision for forecasting the occurrence of objects five
seconds before they appear, averaged over leave-one-out splits. 

\textbf{Baselines:}
We compare against baselines that are similar to our action forecasting
experiments.  However, we add an additional baseline that uses scene features
\cite{zhoulearning} to anticipate objects. One hypothesis is that, since most objects are correlated
with their scene, recognizing the scene may be a good cue for predicting the
onset of objects. We use an SVM trained on state-of-the-art scene features
\cite{zhoulearning}.

\textbf{Results:}
Table \ref{tab:adl} shows average precision for our method versus the baselines
on forecasting objects five seconds into the future.  For the many of the
object categories, our model outperforms the baselines at
anticipating objects, with \textbf{a mean relative gain of 30\%} over baselines. Moreover, our model with
multiple outputs improves over a single output network, suggesting that
handling uncertainty in learning is helpful for objects too. The adapted and off-the-shelf
networks perform similarly to each other in the average. 
Finally, we also qualitatively show some high scoring object predictions in Figure \ref{fig:adl-qual}.

\section{Conclusion}


The capability for machines to anticipate future concepts before they begin is
a key problem in computer vision that will enable many real-world applications.
We believe abundantly available unlabeled videos are an effective resource we
can use to acquire knowledge about the world, which we can use to learn to
anticipate future.




{\small
\textbf{Acknowledgements:} We thank 
members of the MIT vision group for predicting the future on our test set.
We thank TIG for managing our computer cluster, especially Garrett Wollman for troubleshooting
many data storage issues.
We gratefully acknowledge the support of NVIDIA Corporation with the donation of the Tesla K40 GPU used for this research.
This work was supported by NSF grant IIS-1524817, and by a Google faculty research award to AT,
and a Google PhD fellowship to CV.
}

{
\small
\bibliographystyle{ieee}
\bibliography{main}

\begin{thebibliography}{10}\itemsep=-1pt

\bibitem{chen2013watching}
C.-Y. Chen and K.~Grauman.
\newblock Watching unlabeled video helps learn new human actions from very few
  labeled snapshots.
\newblock {\em CVPR}, 2013.

\bibitem{chen2013neil}
X.~Chen, A.~Shrivastava, and A.~Gupta.
\newblock Neil: Extracting visual knowledge from web data.
\newblock {\em ICCV}, 2013.

\bibitem{divvalalearning}
S.~K. Divvala et~al.
\newblock Learning everything about anything: Webly-supervised visual concept
  learning.
\newblock {\em CVPR}, 2014.

\bibitem{doersch2015unsupervised}
C.~Doersch, A.~Gupta, and A.~A. Efros.
\newblock Unsupervised visual representation learning by context prediction.
\newblock {\em ICCV}, 2015.

\bibitem{doersch2012makes}
C.~Doersch, S.~Singh, A.~Gupta, J.~Sivic, and A.~A. Efros.
\newblock What makes paris look like paris?
\newblock {\em ACM Trans. Graph.}, 2012.

\bibitem{donahue2013decaf}
J.~Donahue et~al.
\newblock Decaf: A deep convolutional activation feature for generic visual
  recognition.
\newblock {\em arXiv}, 2013.

\bibitem{fouhey2014predicting}
D.~F. Fouhey and C.~L. Zitnick.
\newblock Predicting object dynamics in scenes.
\newblock {\em CVPR}, 2014.

\bibitem{gong2011multi}
H.~Gong, J.~Sim, M.~Likhachev, and J.~Shi.
\newblock Multi-hypothesis motion planning for visual object tracking.
\newblock {\em CVPR}, 2011.

\bibitem{THUMOS15}
A.~Gorban et~al.
\newblock {THUMOS} challenge: Action recognition with a large number of
  classes, 2015.

\bibitem{hinton2014distilling}
G.~E. Hinton, O.~Vinyals, and J.~Dean.
\newblock Distilling the knowledge in a neural network.
\newblock {\em NIPS}, 2014.

\bibitem{hoai2014max}
M.~Hoai and F.~De~la Torre.
\newblock Max-margin early event detectors.
\newblock {\em IJCV}, 2014.

\bibitem{huang2014action}
D.-A. Huang and K.~M. Kitani.
\newblock Action-reaction: Forecasting the dynamics of human interaction.
\newblock {\em ECCV}, 2014.

\bibitem{ikizler2009learning}
N.~Ikizler-Cinbis, R.~G. Cinbis, and S.~Sclaroff.
\newblock Learning actions from the web.
\newblock {\em ICCV}, 2009.

\bibitem{jia2014caffe}
Y.~Jia, E.~Shelhamer, et~al.
\newblock Caffe: Convolutional architecture for fast feature embedding.
\newblock {\em arXiv}, 2014.

\bibitem{kim2014reconstructing}
G.~Kim and E.~P. Xing.
\newblock Reconstructing storyline graphs for image recommendation from web
  community photos.
\newblock In {\em CVPR}, 2014.

\bibitem{kitani2012activity}
K.~M. Kitani, B.~D. Ziebart, J.~A. Bagnell, and M.~Hebert.
\newblock Activity forecasting.
\newblock {\em ECCV}, 2012.

\bibitem{kooij2014context}
J.~F.~P. Kooij, N.~Schneider, F.~Flohr, and D.~M. Gavrila.
\newblock Context-based pedestrian path prediction.
\newblock {\em ECCV}, 2014.

\bibitem{koppula2013anticipating}
H.~Koppula and A.~Saxena.
\newblock Anticipating human activities using object affordances for reactive
  robotic response.
\newblock {\em RSS}.

\bibitem{krizhevsky2012imagenet}
A.~Krizhevsky et~al.
\newblock Imagenet classification with deep convolutional neural networks.
\newblock {\em NIPS}, 2012.

\bibitem{lampert2015predicting}
C.~H. Lampert.
\newblock Predicting the future behavior of a time-varying probability
  distribution.
\newblock In {\em CVPR}, 2015.

\bibitem{lan2014hierarchical}
T.~Lan, T.-C. Chen, and S.~Savarese.
\newblock A hierarchical representation for future action prediction.
\newblock {\em ECCV}, 2014.

\bibitem{le2013building}
Q.~V. Le et~al.
\newblock Building high-level features using large scale unsupervised learning.
\newblock {\em ICML}, 2013.

\bibitem{lezama2011track}
J.~Lezama et~al.
\newblock Track to the future: Spatio-temporal video segmentation with
  long-range motion cues.
\newblock {\em CVPR}, 2011.

\bibitem{mobahi2009deep}
H.~Mobahi, R.~Collobert, and J.~Weston.
\newblock Deep learning from temporal coherence in video.
\newblock {\em ICML}, 2009.

\bibitem{patron2010high}
A.~Patron-Perez et~al.
\newblock High five: Recognising human interactions in tv shows.
\newblock {\em BMVC}, 2010.

\bibitem{pei2011parsing}
M.~Pei, Y.~Jia, and S.-C. Zhu.
\newblock Parsing video events with goal inference and intent prediction.
\newblock {\em ICCV}, 2011.

\bibitem{arrowoftime}
L.~C. Pickup et~al.
\newblock Seeing the arrow of time.
\newblock {\em CVPR}, 2014.

\bibitem{pintea2014deja}
S.~L. Pintea et~al.
\newblock D{\'e}j{\`a} vu.
\newblock {\em ECCV}, 2014.

\bibitem{pirsiavash2012detecting}
H.~Pirsiavash and D.~Ramanan.
\newblock Detecting activities of daily living in first-person camera views.
\newblock {\em CVPR}, 2012.

\bibitem{ramanan2007leveraging}
D.~Ramanan, S.~Baker, and S.~Kakade.
\newblock Leveraging archival video for building face datasets.
\newblock {\em ICCV}, 2007.

\bibitem{ranzato2014video}
M.~Ranzato et~al.
\newblock Video (language) modeling: a baseline for generative models of
  natural videos.
\newblock {\em arXiv}, 2014.

\bibitem{razavian2014cnn}
A.~S. Razavian et~al.
\newblock Cnn features off-the-shelf: an astounding baseline for recognition.
\newblock {\em arXiv}, 2014.

\bibitem{ryoo2011human}
M.~Ryoo.
\newblock Human activity prediction: Early recognition of ongoing activities
  from streaming videos.
\newblock {\em ICCV}, 2011.

\bibitem{sadanand2012action}
S.~Sadanand and J.~J. Corso.
\newblock Action bank: A high-level representation of activity in video.
\newblock {\em CVPR}, 2012.

\bibitem{srivastava2014dropout}
N.~Srivastava et~al.
\newblock Dropout: A simple way to prevent neural networks from overfitting.
\newblock {\em JMLR}, 2014.

\bibitem{unsupervised}
N.~Srivastava et~al.
\newblock Unsupervised learning of video representations using lstm.
\newblock {\em arXiv}, 2015.

\bibitem{torralba200880}
A.~Torralba et~al.
\newblock 80 million tiny images: A large data set for nonparametric object and
  scene recognition.
\newblock {\em PAMI}, 2008.

\bibitem{vu2014predicting}
T.-H. Vu, C.~Olsson, I.~Laptev, A.~Oliva, and J.~Sivic.
\newblock Predicting actions from static scenes.
\newblock {\em ECCV}, 2014.

\bibitem{walker2014patch}
J.~Walker, A.~Gupta, and M.~Hebert.
\newblock Patch to the future: Unsupervised visual prediction.
\newblock {\em CVPR}, 2014.

\bibitem{walker2015dense}
J.~Walker, A.~Gupta, and M.~Hebert.
\newblock Dense optical flow prediction from a static image.
\newblock {\em arXiv}, 2015.

\bibitem{wang2015unsupervised}
X.~Wang and A.~Gupta.
\newblock Unsupervised learning of visual representations using videos.
\newblock {\em arXiv}, 2015.

\bibitem{xie2013inferring}
D.~Xie, S.~Todorovic, and S.-C. Zhu.
\newblock Inferring" dark matter" and" dark energy" from videos.
\newblock {\em ICCV}, 2013.

\bibitem{yuen2010data}
J.~Yuen and A.~Torralba.
\newblock A data-driven approach for event prediction.
\newblock {\em ECCV}, 2010.

\bibitem{zha2015exploiting}
S.~Zha et~al.
\newblock Exploiting image-trained cnn architectures for unconstrained video
  classification.
\newblock {\em arXiv}, 2015.

\bibitem{zhoulearning}
B.~Zhou et~al.
\newblock Learning deep features for scene recognition using places database.
\newblock {\em NIPS}, 2014.

\end{thebibliography}
}

\end{document}